# Stack Exchange Tagger
(IEE-03: Course Project)

Sanket Mehta
sanketmehta.iitr@gmail.com
Enrolment No: 11114038

Shagun Sodhani
sshagunsodhani@gmail.com.
Enrolment No: 11114039

**Abstract**

The goal of our project is to develop an accurate tagger for questions posted on Stack Exchange. Our problem is an instance of the more general problem of developing accurate classifiers for large scale text datasets. We are tackling the multilabel classification problem where each item (in this case, question) can belong to multiple classes (in this case, tags). We are predicting the tags (or keywords) for a particular Stack Exchange post given only the question text and the title of the post. In the process, we compare the performance of Support Vector Classification (SVC) for different kernel functions, loss function, etc. We found linear SVC with Crammer Singer technique produces best results.

## 1. Main Objectives

- Use SVC with different kernel functions (rbf, linear, polynomial, sigmoid).
- Compare performance with respect to the number of iterations, loss function, regularization term.

## 2. Status and other details

- Fully completed and open sourced. (https://github.com/shagunsodhani/StackExchange-tagger).
- Total time spent on the project: 12 days

## 3. Major stumbling blocks

- *Stack Exchange Dataset*: It took us time to scrape the entire dataset.
- *Computational Power Limitation*: The time complexity for finding Singular Value Decomposition (SVD) for an mxn matrix is $O(m^2n + n^3)$.
- *Choice of Error Metric*: Since multi-label classification is different from multi-class classification, we need to modify accuracy, precision and recall for multi-label classifiers.

## 4. Introduction

Stay organized, get found and promote yourself – 3 reasons why tags are important [1]. Tags are also used as a form of query based search for information retrieval [2]. Tagging of online content by humans is increasing everyday. Hashtags for tweets on Twitter and posts on Facebook and Google Plus are examples of hashtags in social networks. Some work has already been done around this problem to address tag prediction but it still remains a challenge [3]. Facebook also conducted a competiton for predicting tags for questions posted on "Stackoverflow Network". This contest, titled "Facebook Recruiting III - Keyword Extraction" [4], was conducted on Kaggle to recruit developers to Facebook. Our work is also inspired by this contest.

There are many challenges involved in building a tag prediction system to solve this problem. First we need to get data in abundance for training our system. Secondly data should be constrained which means we should have limited number of possible tags. For e.g., in case of Twitter, there is no restriction on hashtags so Twitter dataset is unconstrained in nature. Third real data contains lot of noise so pre-processing of data (Singular Value Decomposition for dimensionality reduction) takes lot of time and is also computationally expensive.

To solve the first two challenges, we used Stack Exchange dataset. Stack Exchange is a network of 130+ Q&A communities including the very popular Stack Overflow, the preeminent site for programmers to find, ask, and answer questions about software development [5]. The Stack Exchange Network covers topics as diverse as Mathematics, Home Improvement, Statistics, English Language and Usage. To overcome computational limitations, we used DELL PRECISION T5600 Sever.

The problem which we are addressing in this paper is an instance of the more general problem of developing accurate classifiers for large scale text datasets (here the dataset comprises of posts made on the StackExchange network). We are tackling the multilabel classification problem where each item (in this case, question) can belong to multiple classes (in this case, tags). We are predicting the tags (or keywords) for a particular Stack Exchange post given only the question text and the title of the post.

Given the text and the title, we first parse the data to get rid of stop-words. Next we perform stemming and lemmatiztion. This is followed by tf-idf based filtering and then we extract features using SVD. Once we have our training data in form of features and classes, we train various classifiers with linear, polynomial, sigmoid and rbf kernels. We vary the number of iterations and the error function as well and do a comprehensive comparison of the different approaches for different values of the parameters.

The organization of the paper is as follows. Section 5 summarises related work in this field. Section 6 deals with the proposed approach. It also deals with the feature vector extraction mechanism and dimensionality reduction. Section 7 presents the results of our experiments. Section 8 concludes the paper and section 9 recommends directions for future extension of our work.

## 5. Related Work

[3] focuses on mining user interest from their behavior on stackoverflow.com and leveraging that information for predicting tags. Also they focus only on stackoverflow.com and not other member sites of the StackExchange network. Our work is different from existing work as none of the existing work does a survey analysis. Also most of the related work focus on getting good results for a given member site of Stack Exchange Network while in our case, we keep all the methods to be very generalized thereby making them applicale in all the member sites. [10] uses a co-occurrence model that predicts tags based on the words in the post and their relation (co-occurrence) to tags. They built model for StackOverflow dataset by constraining the next word predicted to only tags. His co-occurrence model has a 47% classification accuracy predicting one tag per post. Our experimental results show that we beat his accuracy as mentioned in Section 7.

## 6. Proposed Approach

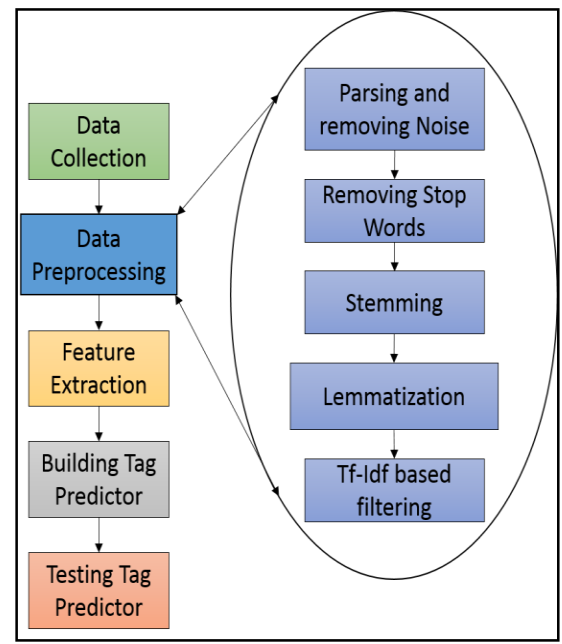

**Figure 6.1 Proposed System**

Figure 6.1 shows proposed workflow of our system. We explain each step in detail in following subsection.

### 6.1 Data Collection – Stack Exchange Data

StackExchange Network provides all community-contributed content under the Creative Commons BY-SA 3.0 license. A quarterly dump of all this data (after sanitization) is updated on the Internet Archive. Other than this method, all the data ia accessible via StackExchange API. We have used both the dumps as well as the API to get our data. This data included information about Posts, Users, Votes, Comments, Badges, PostHistory, and PostLinks. Of these, we kept the information related to the problem and tags and filtered out the remaining information. Figure 6.2 shows snapshot of a example from stackoverflow.com member site.

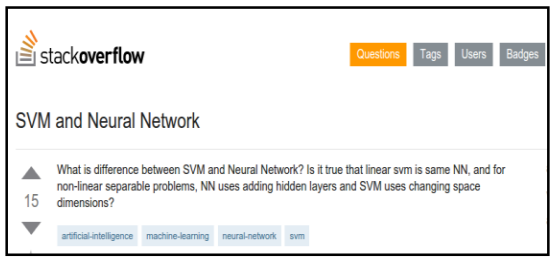

**Figure 6.2 Snapshot of a example from stackoverflow.com**

## 6.2 Data Preprocessing

### 6.2.1 Parsing and Removing Noise

The content obtained from Stack Exchange archives and by scraping is in html format. So we first parse out the text part by filtering HTML tags. Next we remove any code snippets that users might have added with their question and retain only the words used in the question itself.

### 6.2.2 Removing Stop Words

Stop words refer to words like "and", "or", "the" etc which do not add any specific information about the context of text. These words are normally removed as a part of preprocessing stage. There is no single universal list of stop-words which can be used in all contexts. In many cases, developers have to come up with their own list of stopwords. Also what is stopword in one context, may not be stopword in another context. Eg we may normally treat mathematical symbols as stopwords but they become relevant if our text contains words like C++.

### 6.2.3 Stemming

Stemming [6], [7] refers to the process of reducing words to their word root, also called as word stem, and hence the name stemming. A program that can perform stemming is referred to as stemmer. E.g., words "fishing", "fished", and "fisher" would be stemmed to the word "fish". Most Information Retrival systems use stemming as a preprocessing step before storing data or before performing applying more sophisticated techniques on user data. A lot of algorithms are available for stemming. The prominent ones include the porter stemmer, the snowball stemmer and the lancaster stemmer. Porter stemmer is the most comman algorithm and consists of 5 phases of word reduction that are applied sequentially.

We have used porter stemmer [8] in our implementation as well.

### 6.2.4 Lemmatization

Lemmatization is the process of grouping together different forms of a word so as to treat them as a single word. This single word is called lemma and hence the name lemmatization. E.g., the verb 'to eat' may appear as 'eat', 'ate', 'eating', etc though all these words can be reduced to a common lemma i.e., 'eat'.

We have used the 'Wordnet lemmetizer' in our implementation.

### 6.2.5 Tf-Idf based filtering

tf–idf [9] (term frequency–inverse document frequency) is defined for a word given a collection of documents (also called a corpus). It indicates how important the word is for the given corpus. We have used it as a weighing factor to remove some words that do not convey information about the context of the problem at hand. The importance varies proportionally with the number of times the word appears in the document and is inversely proportional to the frequency of the word in the corpus.

$$tf(t,d) = 0.5 + \frac{0.5 * f(t,d)}{\max\{f(w,d): w \in d\}}$$

Where, $t$ refers to term,
$d$ refers to document,

$tf(t,d)$ is term frequency,
$f(t,d)$ is the raw frequency of a term in a document.

$$idf(t,D) = \log \frac{N}{|\{d \in D : t \in d\}|}$$

Where, $N$ is the total no. of documents in the corpus,
$|\{d \in D : t \in d\}|$ is number of documents where the term $t$ appears. Finally tf-idf is calculated as:

$$tfidf(t,d,D) = tf(t,d) * idf(t,D)$$

### 6.3 Feature Extraction

Feature extraction refers to the process of deriving features/values from the given dataset such that the derived features are more informative and less redundant than the parameters in the given dataset. This is closely related to dimensionality reduction where in we reduce the number of dimensions of the given dataset to make computations feasible. Some important techniques include SVD (Singular Value Decomposition) and PCA (Principle Component Analysis). We have used SVD and will be explaining it further.

SVD [9] is a dimensionality reduction technique that produces a factorization of any matrix, real or complex. SVD connects the rows and columns of a matrix by defining a small number of "concepts".

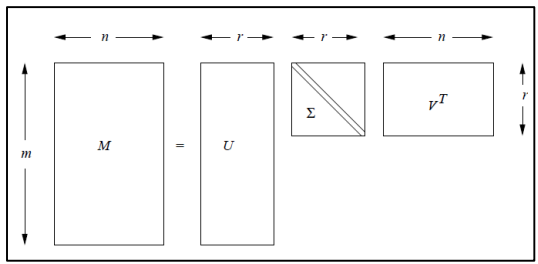

**Figure 6.3 The form of a Singular Value Decomposition(Courtesy [9])**

Let $M$ be an m × n matrix, and let the rank of $M$ be r. Rank of a matrix is the largest number of rows (or equivalently columns) that we can choose for which no nonzero linear combination of the rows is the all-zero vector 0. Figure 6.3 shows the form of a Singular Value Decomposition. Then, given $M$, we can find matrices $U$, $S$, and $V$ such that:

$$M = U\Sigma V^T$$

Where, $U$, $\Sigma$ and $V^T$ satisfies the following properties:
1. $U$ is an m x r column-orthonormal matrix.
2. $V$ is an n x r column-orthonormal matrix.
3. $\Sigma$ is a diagonal matrix.

The diagonal entries $\sigma_i$ of $\Sigma$ are known as the singular values of $M$. If we list the singular values in descending order, the diagonal matrix $\Sigma$ is uniquely determined by $M$.

### 6.4 Building Tag Predictor

#### 6.4.1 Support Vector Classification (SVC)

We consider one-vs-all classifier. Given training vectors $x_i \in \mathbb{R}^p, i = 1, \ldots, n$ in 2 classes, and a vector $y_i \in \{1, -1\}^n$, our primal problem formulation is as follows:

$$\min \frac{1}{2} w^T w + C \sum_{i=1}^{n} \xi_i$$

subject to
$$y_i(w^T \phi(x_i) + b) \geq 1 - \xi_i, \xi_i \geq 0, i = 1, \ldots, n$$

Its dual is as follows:
$$\min \frac{1}{2} \alpha^T Q \alpha - e^T \alpha$$

subject to
$$y^T \alpha = 0 \text{ and } 0 \leq \alpha_i \leq C, i = 1, \ldots, n$$

where, $e$ is the vector of all ones,
$C > 0$ is the upper bound and $C$ is regularization parameter,
$Q$ is an $n$ by $n$ positive semidefinite matrix,
$Q_{ij} = K(x_i, x_j) = \phi(x_i)^T \phi(x_j)$ is the kernel.

The decision function as defined in [11], [13] is:
$$sgn(\sum_{i=1}^{n} y_i \alpha_i K(x_i, x) + \rho),$$

Where, $\rho$ is intercept.

We have considered various kernel functions in our case – rbf, linear, polynomial with degree = 2 and 3 and also sigmoid. Comparative study of all these kernels is presented in the next section. We also varied $C$. Large $C$ means we are modeling hard-margin svc which leads to low training error but poor generalization. We also vary the

### 6.4.2 Linear Support Vector Classification (Linear SVC)

Linear SVC is SVC with a Linear Kernel. We are performing further tweaking with linear SVC as our previous results indicated that Linear SVC peforms better than SVC with other kernels. When using Linear SVC, we experiment around with both the loss function and with the optimization technique - namely the traditional multi-class optimization technique or the crammer singer approach. We played around with "hinge" loss function and "squared hinge" loss function. Next we take up the traditional multi-class optimization technique vs crammer singer approach.

The primary approach for solving multiclass problems using support vector machines has focused on reducing a single multiclass problems into multiple binary problems. For e.g., we may build a set of binary classifiers to distinguish between labels. This approach is more commonly known as the one-vs-rest approach. An alternate method was proposed by Crammer and Singer [12]. They have used the dual of the optimization problem to incorporate kernels with a compact set of constraints and decomposed the dual problem into multiple optimization problems, each of reduced size. They then use a fixed-point algorithm to solve these reduced optimization problems. This way crammer singer approach optimizes a joint objective over all classes. Also in crammer singer approach, the results are not affected by the loss function used which we infer from the next section.

### 6.5 Testing Tag Predictor

Multi-label classification is different from multi-class classification and hence requires different metrics than the ones we use for traditional multi-class classification. The error metrics that we have used are proposed in [14] for multi-label classification problems.

Let $D$ be a multi-label evaluation data set, consisting of $|D|$ multi-label examples $(x_i, Y_i), i = 1..|D|, Y_i \subseteq L$. Let $H$ be a multi-label classifier and $Z_i = H(x_i)$ be the set of labels predicted by $H$ for $x_i$. The following metrics for the evaluation of $H$ and $D$ are used:

$$Accuracy(H,D) = \frac{1}{|D|} \sum_{i=1}^{|D|} \frac{|Y_i \cap Z_i|}{|Y_i \cup Z_i|}$$

$$Precision(H,D) = \frac{1}{|D|} \sum_{i=1}^{|D|} \frac{|Y_i \cap Z_i|}{|Z_i|}$$

$$Recall(H,D) = \frac{1}{|D|} \sum_{i=1}^{|D|} \frac{|Y_i \cap Z_i|}{|Y_i|}$$

We define percentage error as follows:
$$error = \big(1 - Accuracy(H,D)\big) * 100.$$

### 7. Experimental Results

We considered total 10,000 questions. We divided the preprocessed data into training set (80%) and testing set (20%). We applied k-fold cross validation to obtain an average accuracy. We retain 90% of variance when using SVD. The number of features after applying SVD are ~3,000. The number of classes we are dealing with are 10. All the results below are for exact matching (as opposed to atleast one match). All the code used in these experiment has been implemented from scratch and has been open sourced on github [15].

**Table 7.1 Training Errors for linear SVM. (Variation with change in penalty term and number of iterations)**

| SVC (Kernel = RBF) | C = 1000 (hard) | C = 0.001 (soft) |
|---|---|---|
| 200 | 39.0 % | 47.0 % |
| 400 | 31.1 % | 62.6 % |
| 600 | 23.4 % | 36.1 % |
| 800 | 22.62 % | 36.1 % |
| 1000 | 22.34 % | 36.1 % |

Table 7.1 shows the performance of SVC with RBF kernel for the training dataset for the soft margin and the hard margin case as the number of iterations are varied. As the number of iterations increases, the training error decreases. Also soft-margin has more training error which means good generalization as expected.

**Table 7.2 Testing Errors for linear SVM. (Variation with change in penalty term and number of iterations)**

| SVC (Kernel = RBF) | C = 1000 (hard) | C = 0.001 (soft) |
|---|---|---|
| 200 | 54.5 % | 54.87 % |
| 400 | 50.0 % | 66.13 % |
| 600 | 43.6 % | 48.5 % |
| 800 | 43.5 % | 48.5 % |
| 1000 | 43.2 % | 48.5 % |

Table 7.2 shows the performance of SVC with RBF kernel for the testing dataset for the soft margin and the hard margin case as the number of iterations are varied. As the number of iterations increases, the testing error decreases. Also rbf kernel is able to beat the method in [10].

**Table 7.3 Training Error for SVC. (Variation with change in penalty term and kernel)**

| Kernel | C = 1000 (hard) | C = 0.001 (soft) |
|---|---|---|
| RBF | 21.8 % | 36.1 % |
| Linear | 19.0 % | 29.5 % |
| Polynomial (n=2) | 24.3 % | 31.1 % |
| Polynomial (n=3) | 34.0 % | 83.2 % |
| Sigmoid | 83.2 % | 83.2 % |

Table 7.2 shows the performance of SVC with RBF kernel for the testing dataset for the soft margin and the hard margin case as the number of iterations are varied. As the number of iterations increases, the testing error decreases. Also rbf kernel is able to beat the method in [10].

Table 7.3 shows the performance of SVC with different kernel function for the training dataset for the soft margin and the hard margin case while the number of iterations fixed to 10,000. As we can infer that linear kernel performs best followed by rbf then polynomial with degree 2 and polynomial of degree 3. Sigmoid kernel gives the worst performance. Also soft-margin has more training error which means good generalization as expected.

**Table 7.4 Testing Error for SVC. (Variation with change in penalty term and kernel)**

| Kernel | C = 1000 (hard) | C = 0.001 (soft) |
|---|---|---|
| RBF | 43.1 % | 48.5 % |
| Linear | 51.9 % | 45.2 % |
| Polynomial (n=2) | 54.4 % | 65 % |
| Polynomial (n=3) | 72.2 % | 84.4 % |
| Sigmoid | 84.4 % | 84.4 % |

Table 7.4 shows the performance of SVC with different kernel function for the testing dataset for the soft margin and the hard margin case while the number of iterations fixed to 10,000. As we can infer that linear kernel performs best (soft-margin) followed by rbf then polynomial with degree 2 and polynomial of degree 3. Sigmoid kernel gives the worst performance. Also soft-margin has less testing error which means good generalization as expected.

**Table 7.5 Training Error for Linear SVC. (Variation with change in error function and technique)**

| Technique | Hinge Loss Function | Square Hinge Loss Function |
|---|---|---|
| One-vs-rest | 37.52 % | 67.79 % |
| Crammer Singer | 30.71 % | 30.71 % |

Table 7.5 shows the performance of linear SVC with different error functions and techniques. C is set to 0.001 (soft-margin) and the number of iterations is fixed to 10,000. First we observe that training error remains same for Crammer Singer technique irrespective of the error function. Crammer Singer technique performs better than One-vs-rest approach. For One-vs-rest, Square Hinge Loss function gives more training error because outliers are penalized more.

**Table 7.6 Testing Error for Linear SVC. (Variation with change in error function and technique)**

| Technique | Hinge Loss Function | Square Hinge Loss Function |
|---|---|---|
| One-vs-rest | 47.59 % | 68 % |
| Crammer Singer | 45.25 % | 45.25 % |

Table 7.6 shows the performance of linear SVC with different error functions and techniques. C is set to 0.001 (soft-margin) and the number of iterations is fixed to 10,000. First we observe that testing error remains same for Crammer Singer technique irrespective of the error function. Crammer Singer technique performs better than One-vs-rest approach. For One-vs-rest, Square Hinge Loss function gives more testing error because outliers are penalized more.

## 8. Conclusion

We conclude that linear SVC performs better than all other kernel functions in case of both soft and hard margin problem. In case of linear SVC, linear SVC with Crammer Singer technique for soft-margin performs better than ome-vs-rest technique. The best accuracy obtained in our case is 54.75%.

## 9. Future Scope

Feature selection (dimensionality reduction) is a computationally expensive step, so we need to deal with this step for large data size. Also for our analysis we considered only the text part of the data and ignored any code segements or user information present in the system. Also many tags co-occur. E.g., a question tagged "android" would likely be tagged "java" as well. We did not try to learn these co-occurences. These considerations can help to further improve upon accuracy.